\documentclass[twocolumn,10pt]{asme2e}
\pdfoutput=1
\usepackage{graphicx}
\usepackage{subfigure}
\usepackage{gensymb}
\usepackage{float}
\usepackage{multirow}
\usepackage{graphicx}
\usepackage[dvipsnames]{xcolor}
\usepackage[mathscr]{euscript}

\confshortname{MSEC 2023}
\conffullname{the ASME 2023 18th International \\ Manufacturing Science and Engineering Conference}
\confdate{June 12 - June 16}
\confyear{2023}
\confcity{New Brunswick, NJ}
\confcountry{USA}

\papernum{MSEC2023-104127}

\title{Explainable AI for Tool Wear Prediction in Turning}

\author{Saleh Valizadeh Sotubadi
    \affiliation{
	Department of Mechanical \\ Engineering - Engineering Mechanics \\
	Michigan Technological University\\
	Houghton, Michigan 49931\\
    Email: svalizad@mtu.edu
    }	
}

\author{Rui Liu
    \affiliation{
    Department of Mechanical Engineering\\
	Kate Gleason College of Engineering\\
	Rochester Institute of Technology\\
	Rochester, New York 14623\\
	Email: rleme@rit.edu
    }
}

\author{Vinh Nguyen\thanks{Corresponding Author.}
    \affiliation{
	Department of Mechanical \\ Engineering - Engineering Mechanics \\
	Michigan Technological University\\
	Houghton, Michigan 49931\\
    Email: vinhn@mtu.edu
    }	
}

\begin{document}

\maketitle    

\begin{abstract}
{\it This research aims develop an Explainable Artificial Intelligence (XAI) framework to facilitate human-understandable solutions for tool wear prediction during turning. A random forest algorithm was used as the supervised Machine Learning (ML) classifier for training and binary classification using acceleration, acoustics, temperature, and spindle speed during the orthogonal tube turning process as input features. The ML classifier was used to predict the condition of the tool after the cutting process, which was determined in a binary class form indicating if the cutting tool was available or failed. After the training process, the Shapley criterion was used to explain the predictions of the trained ML classifier. Specifically, the significance of each input feature in the decision-making and classification was identified to explain the reasoning of the ML classifier predictions. After implementing the Shapley criterion on all testing datasets, the tool temperature was identified as the most significant feature in determining the classification of available versus failed cutting tools. Hence, this research demonstrates capability of XAI to provide machining operators the ability to diagnose and understand complex ML classifiers in prediction of tool wear.}
\end{abstract}
\begin{nomenclature}
\entry{$g(x^{'})$}{Explanation Model}
\entry{$M$}{Number of Input Features}
\entry{$Acc$}{ML Model Accuracy}
\entry{$a_X$}{Tool Acceleration Along the x Axes}
\entry{$a_Y$}{Tool Acceleration Along the y Axes}
\entry{$a_Z$}{Tool Acceleration Along the z Axes}
\entry{$s_1$}{Acoustic Data Collected by the MEMS Microphone}
\entry{$s_2$}{Acoustic Data Collected by the Electret Microphone}
\entry{$\theta$}{Temperature Data Collected by the Thermometer}
\entry{$\omega$}{Spindle Speed}
\entry{$TP$}{ True Positive}
\entry{$FP$}{ False Positive}
\entry{$TN$}{ True Negative}
\entry{$FN$}{ False Negative}
\entry{$TPR$}{True Positive Rate}
\entry{$FPR$}{False Positive Rate}
\entry{$ROC$}{Receiver Operating Characteristic}
\entry{$MCC$}{Matthews Correlation Coefficient}
\end{nomenclature}
\section{INTRODUCTION}

Machining is one of vital manufacturing processes to convert engineering designs into real-world objects. For a machining process to be successful, several factors have to be considered. Specifically, tool wear is a critical factor that can result in negative effects on the machining process in many aspects, including severe plastic deformation, mechanical breakage, cutting edge blunting, high cutting temperature, and low cutting efficiency \cite{Tool_Wear}. 

Due to its significant effects on the machining quality, several studies have been conducted for tool wear monitoring. In \cite{Tool_Wear_1}, a model-based online approach was developed to monitor tool wear during the machining process. In \cite{Tool_Wear_2}, an online sensing method was proposed to monitor the tool wear, and a feedback control method was developed to compensate for the dimension errors by constricting the tool and the workpiece to the designated tolerance zone. In \cite{Tool_Wear_3}, an in-process tool wear methodology was developed to monitor the tool wear in turning based on cutting power measurements. 

Manufacturing has been benefited from the emerging growth of Machine Learning (ML) applications \cite{ML_manufacturing, ML_manufacturing_2, ML_Manufacturing_3, ML_Manufacturing_4}. Specifically for tool wear analysis, several ML algorithms have been applied by prior studies in analyzing and predicting the tool wear \cite{Toolwear_ML}. In \cite{ToolWear_ML_2}, an Artificial Neural Network (ANN) was adopted to predict the surface roughness and tool wear in turning processes, and the results showed that the ANN was capable of predicting the tool wear with an reasonable accuracy. In \cite{ToolWear_ML_03}, a multisensor fusion model was developed to predict flank tool wear in turning processes. For this purpose, an ANN and a regression model were trained. The two models would merge multiple input features including the cutting force, cutting temperature, and vibration signals for the training and prediction tasks. The results showed that the developed framework was able to predict the tool wear with higher accuracy. In \cite{ToolWear_ML_4}, a Support Vector Machine (SVM) algorithm was implemented to train a ML framework for tool breakage detection in milling process, and two factors, including the cutting force and the power consumption, were considered as input features for the training and testing tasks of the developed ML framework. 

All the aforementioned studies demonstrated the successful integration of ML algorithms with various/multiple sensors in monitoring the tool wear during the machining process. However, most of those studies lack an explainable framework to describe the logic behind decisions made by the ML model, especially the level of contribution of each input feature in the final decision making of the ML model. A framework for tasks including tool wear prediction would be beneficial for process engineers with valuable information to better understand the logic behind ML model predictions. Also, such a framework can aid in understanding the most contributing factors to mitigate the effects of tool wear by adjusting critical process parameters. 

This study aims to resolve the aforementioned issue by proposing an Explainable Artificial Intelligence (XAI) framework for tool wear prediction utilizing any ML methodology. For this purpose, an ensemble learning ML model was developed and trained on a set of training data acquired from orthogonal turning experiments. Subsequently, the trained model was evaluated based on a game theory-based tabular explanation algorithm to describe the logic behind decisions made by the ML model. It is expected that the developed XAI framework is capable of reasonably explaining the trained ML model and provide useful information for tool wear prediction. 
 
\section{METHODOLOGY}
\subsection{Machine Learning Framework: Random-Forest Algorithm}
The ML framework employed in this work is a supervised classifier developed for binary classification purposes, which has been used in distinguishing worn tools from their unworn counterparts. The Random-Forest classification (RFC) model is used to map a set of input features $X$ to their corresponding and known labels $Y$, which is an ensemble learning technique comprised of a collection of decision trees and has been demonstrated for binary as well as multi-class classification tasks \cite{RFC_1, RFC_2}. RFC is computationally efficient, robust to noise, and able to adapt the nonlinear patterns between input and output data. When an input is introduced to the classifier, it propagates among all the decision trees, and each tree assigns a unit vote predicting the corresponding class of the input data. The overall prediction of the RFC is the class having the highest number of votes. 

In this study, the ML framework was developed using Scikit-learn package in the Python programming language and trained upon the pre-processed training data collected from the experiments. After the training process, the trained model was evaluated using testing data input features. For this purpose, multiple accuracy metrics were implemented for model evaluation. The results of these evaluation metrics will be discussed in more detail in the subsequent sections. 

\subsection{Shapley Criterion}
The main contribution of this study is to develop an Explainable AI (XAI) framework to explain the predictions of ML models used in tool wear prediction. Specifically, ML stakeholders are interested in realizing the relative contribution of input features on the final decision of the ML model for a given input. For this purpose, the SHapley Additive exPlanations (SHAP) criterion stemed from game theory was utilized in this study for tabular data \cite{Shap_2}. Through the SHAP criterion, the prediction of the model is decomposed among all the input features involved in the decision making process. This decomposition is achieved by additive feature attribution analysis described as follows.

\begin{equation} \label{eq_2_1}
    g(x^{'}) = \phi_{0} +  \sum_{i=1}^{M} \phi_{i} x_{i} ^ {'}
\end{equation}

\begin{equation} \label{eq_2_2}
    \phi_{i} = \sum_{S \subseteq N/i}^{} \frac{|S|!(M - |S| - 1)!}{M!}[F_{X}(S \cup i) - F_{X}(S)] 
\end{equation}

Here, $g(x^{'})$ is the explanation model, where $x^{'} \in \{0, 1\}^{M}$. $M$ is the number of features, and $\phi_{i} \in R$. Eq (\ref{eq_2_2}) describes how the SHAP criterion determines the contribution of all the input features in the model predictions. Using this method, the model is trained on all the feature subsets $S \subseteq F$, where $F$ represents the set of all features. To calculate the contribution of each input feature, the model $F_X(S \cup {i})$ is trained by including the feature $i$. Subsequently, model $F_X(S)$ is trained excluding the feature $i$ from the input features. When the two models are trained, the predictions of a specific input $F_{X}$ on the two models are compared together. The differences of the results between two models are the indications of each input feature's effect on the overall decision of the ML model \cite{Shap_1}. 

\section{EXPERIMENTAL SETUP AND DATA ACQUISITION}
\subsection{Experimental Setup}
\begin{figure}[t]
    \hspace*{-0.5cm}
    \includegraphics[width=0.5\textwidth]{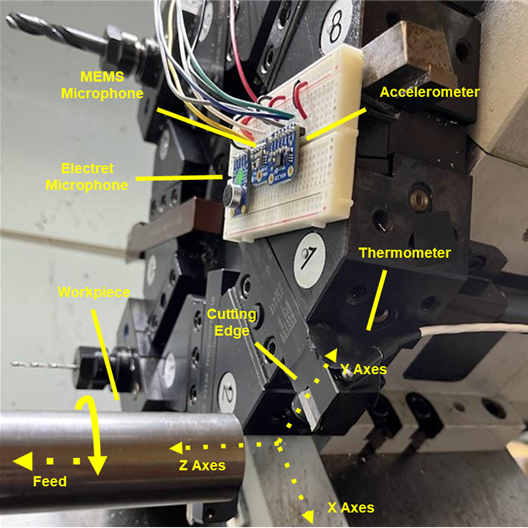}
    \caption{Experimental setup for orthogonal turning experiments.}
    \label{fig_setup}
\end{figure}

Figure \ref{fig_setup} shows the experimental setup used in the orthogonal cutting experiments. All cutting tests were performed by turning 1018 steel tubes with a wall thickness of 2.11 mm and a constrained stick-out of 76.2 mm on a 2-axis lathe (15-L Slant Pro, Tormach). The uncoated tungsten carbide inserts (1691N82, McMaster-Carr) were mounted into a 19.05 mm square tool holder with 5\degree rank angle (3288A851, McMaster-Carr). Cuts were conducted over an orthogonal cutting distance of 11.48 m with a feed of 0.025 mm and  three cutting speeds of 54 m/sec, 69 m/sec, and 84 m/sec. During the experiments, time series data were collected from (1) a 20 Hz - 20 KHz electret microphone (MAX9814, Maxim Integrated), (2) a 100 Hz-10 KHz MEMS-based microphone (SPW2430, Knowles), (3) a 3g 3-axis accelerometer (ADXL335, Analog Devices), and (4) a digital thermometer (DS18B20, Maxim Integrated). The measurements were recorded by a 32KB flash, 2KB, 16 MHz clock micro-controller (METRO 328, Adafruit). The data were recorded at approximately 38 ms and then written to an SD card at the end of the experiments (Data Logging Shield for Arduino, Adafruit).

\begin{figure}[t]
    \hspace*{-0.5cm}
    \includegraphics[width=0.5\textwidth]{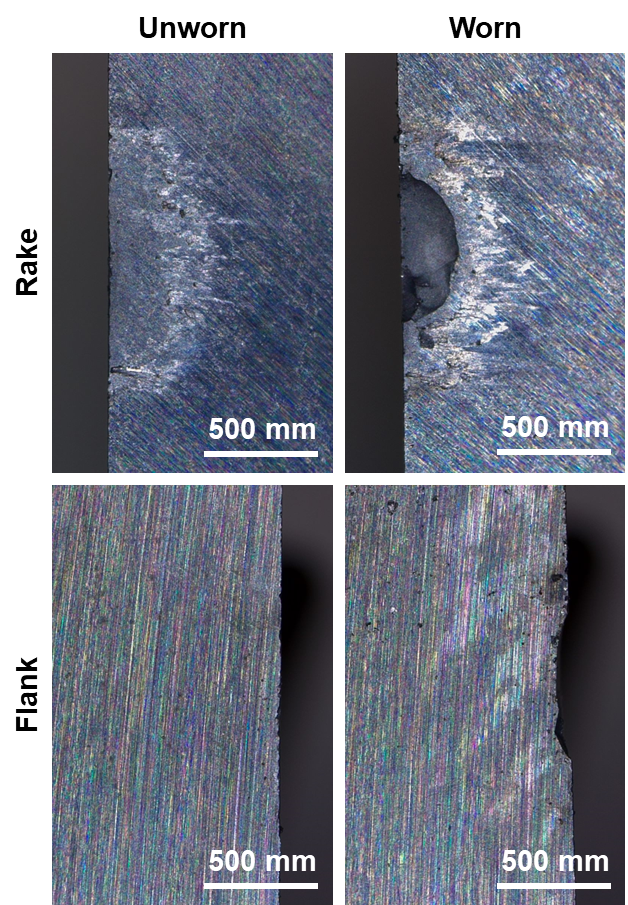}
    \caption{Optical microscope wear images at 69 m/sec.}
    \label{fig_wear}
\end{figure}

After each cut, the rake and flank surfaces were imaged on a stereoscopic optical microscope (DVM6, Leica). Figure \ref{fig_wear} shows example tool wear images at 69 m/sec. Figure \ref{fig_wear} shows that as the tool is unworn after a cut, a tool-chip contact area is formed but the integrity of the tool is conserved. However, after the tool is worn due to repeated cutting, the cutting edge becomes fractured and breaks from the cutting tool. Thus, time series data corresponding to a worn tool according to the optical images was labeled as worn for classification. {It is also worthwhile to note that the occurrence of chipping was the main criterion to determine the tool as worn or unworn. When the first worn incident was observed at each spindle speed, the first couple of signal windows corresponding to that specific observation were discarded and only the last two data windows were considered for the training process of the ML model. This is because the initial data windows relating to that specific cutting process might not have represented that worn tool. Specifically, during the initial seconds of the cutting process, the tool might have experienced a transition from the unworn station to the worn condition. Therefore, to be more accurate and avoid any confusion during the training process of the ML model, the first couple of signal windows representing the first worn observation were not considered within the training/testing data.}

\subsection{Data Acquisition and Preprocessing}

\begin{figure}[t]
     \centering
     \subfigure[Acquired microphone data through the MEMS microphone during the experiment with a spindle speed of 700 rpm]{\includegraphics[width=8cm,height=4.5cm]{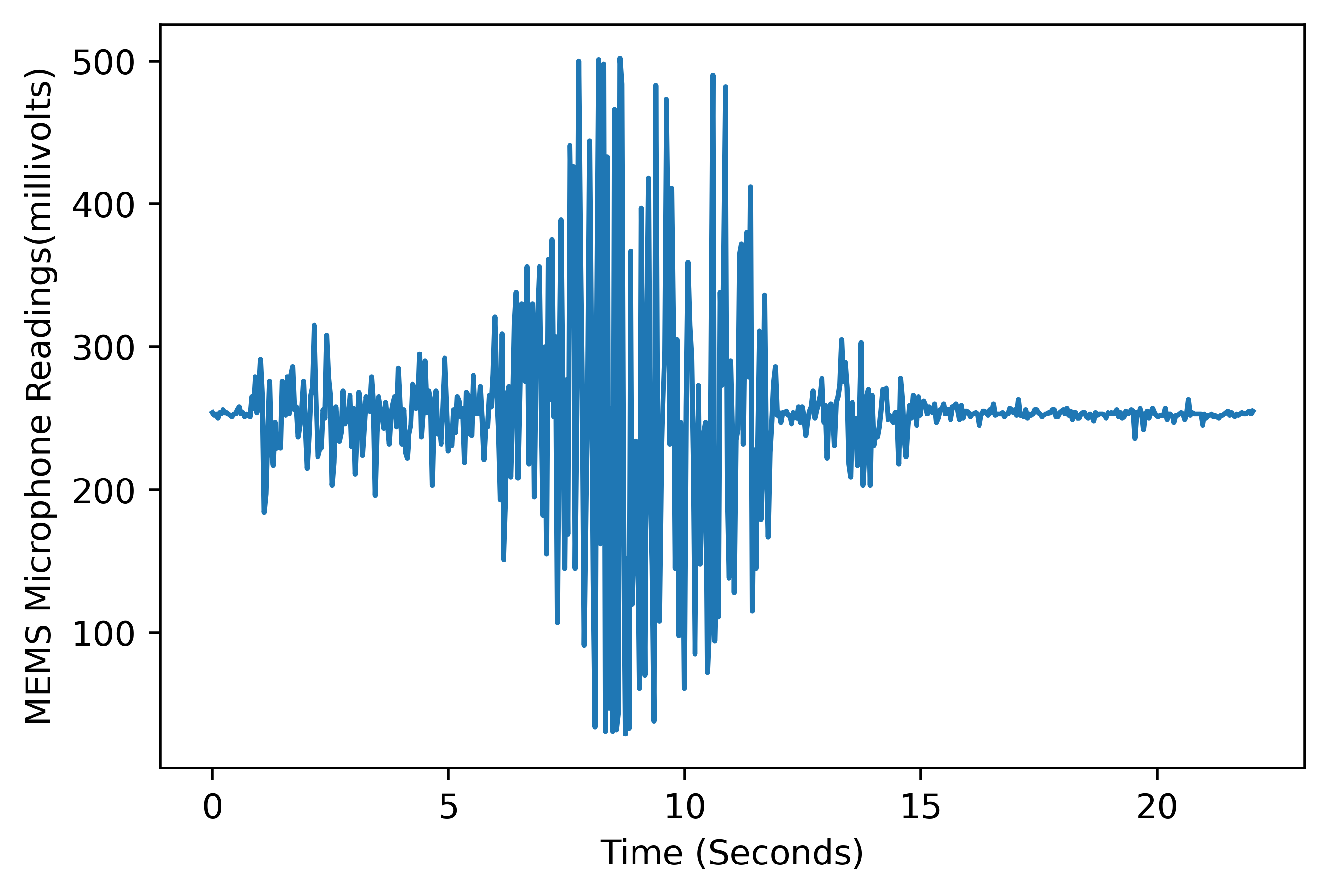}}
    \hfill
    \hspace*{-0.35cm}
    \subfigure[Recorded acceleration of the tool along the vertical axes through the accelerometer module during the experiment with a spindle speed of 700 rpm ]{\includegraphics[width=8cm,height=4.5cm]{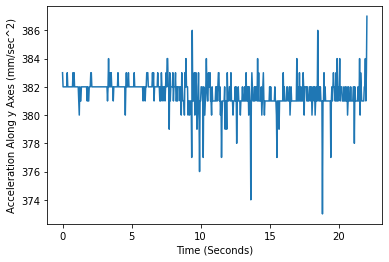}}
    \hfill
    \hspace*{-0.35cm}
    \subfigure[Temperature data collected by the thermometer during the experiment with a spindle speed of 700 rpm]{\includegraphics[width=8cm,height=4.5cm]{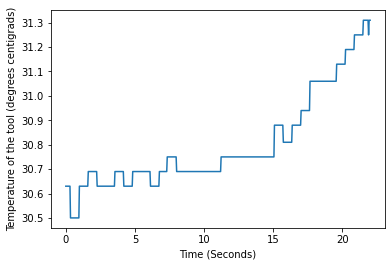}}
    \caption{Time domain representation of a set of acquired input data during the experiments}
    \label{fig_raw_data}
\end{figure}

Corresponding to the three tested cutting speeds, the spindle speeds of 700, 900, 1100 rpm during each cut were recorded as an input feature. In addition to spindle speed, other sets of data were acquired using the sensors mounted on experimental setup as shown in Fig. \ref{fig_setup}. The collected sensor data were the recorded three dimensional acceleration information, the acoustics data through different microphones, and the recorded thermal data through the thermometer sensor. {It should be stated that the intention for using accelerometer sensors relies on the reasons stated as follows: (1) it is assumed that the cutting piece itself is inhomogeneous, and (2) the eccentricity of the cutting piece held by the chuck relative to the cutting tool. These are two reasons resulting in uneven depths of cut that correspond to uneven force distributions between the cutting piece and cutting tool during the cutting process. Hence, this variation induces acceleration during the cutting process. Additionally \cite{ACC_1, ACC_2, ACC_3}, have used accelerometers during the cutting process to assess machining parameters in their research.} After all the cutting operations were completed, the sensory data representing the unprocessed input data were collected. Figure \ref{fig_raw_data} depicts the raw acoustic data, acceleration data, and temperature data with the spindle speed set to 700 rpm in the time domain acquired by the MEMS microphone, the accelerometer module, and the thermometer, respectively. The collected sensor data were then pre-processed and used as a tuple of input features for ML model training and testing. Since the collected data were temporal, each time series of data was windowed and divided into seven different windows with equal number of data points. Because temporal data cannot be used as an input to the RFC model, the variance of acceleration and acoustic time series was calculated, while the area under the curve over the time span of the temperature data was calculated. The tuple storing different features of any input data would be expressed as $X_i = (a_X, a_Y, a_Z, s_1, s_2, \theta, \omega)$, where $X_i$ indicates the $i^{th}$ input. $a_X$, $a_Y$, and $a_Z$ are the acceleration information of the cutting tool during the cutting process along the Cartesian axes. $s_1$, and $s_2$ are the acoustic data collected from two different microphones, $\theta$ represents the temperature data collected from the thermometer, and $\omega$ is the spindle-speed that is predefined by the operator.
As mentioned in the previous part, the class of each experimental run was determined after the experiment was completed. For this purpose, surface of the tool was observed through imaging techniques and at different angles, and the condition of the tool was categorized as either unworn, or worn.


\section{RESULTS AND DISCUSSION}
{In this research, a total of 23 experiments were carried out at three different cutting speeds. After the preprocessing was done on all the collected datasets, each data was equally divided into seven time windows expect for the incidents where the worn tool was observed for the first time in which the first five sets of data were excluded from the data to be more accurate on segregating the worn tool from its unworn counterpart. Therefore, there were a total of 146 data points, including training and test datasets, extracted and used for the training process of the ML model.} The data was divided into training and test sets. {In this research, no independent validation data was used during the training of the ML model. This is because the independent validation dataset is mostly used to assess the ML model during the training and avoid over-fitting in the ML model while the training process is ongoing. However, the ML model in this model did not suffer from over-fitting because of sufficient datasets and optimal architectural design of the ML model.} The training data comprised of $60\%$ of the entire dataset, while the remaining $40\%$ was constituent of the test dataset. The model was trained on a computer (Intel Core i5-10600 processor, 16 Gbs of RAM). After the model training, the test set was used by the ML model for prediction. Furthermore, the Shapley criterion was implemented to describe the overall performance of the ML model.
\subsection{Performance of the Machine Learning Model}

\begin{figure}[t]
    \hspace*{-0.5cm}
    \includegraphics[width=0.5\textwidth]{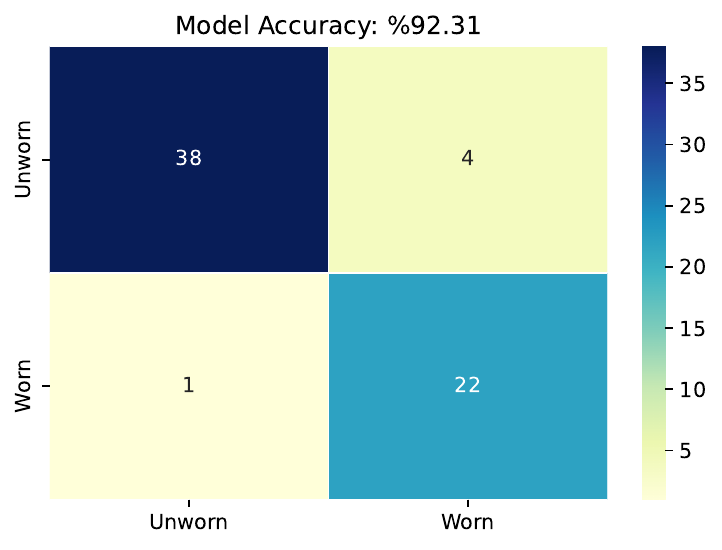}
    \caption{Confusion Matrix describing the overall performance of the random forest algorithm on the test data, after the training process was executed}
    \label{fig_4_1}
\end{figure}

Figure \ref{fig_4_1} illustrates the confusion matrix of the developed RFC on the test data. The horizontal axis indicates the real class of each of the test data, whereas, the vertical axis represents the outcome of the ML model. Since the classification task is a binary classification, the predictions of the model were divided into four categories, which are referred to as True Positive ($TP$), True Negative ($TN$), False Positive ($FP$), and False Negative ($FN$). In this study $TP$, and $TN$ occurred when the model correctly determined whether the tool is unworn or worn, respectively. On the contrary, cases occurred where the model was incorrect by deciding whether the tool was unworn ($FP$), or worn ($FN$). To quantify the model performance, four metrics were considered in this study in addition to qualitative representations. The accuracy value of the model is presented in Table \ref{tab_4_1} along with other metric factors as follows.

\begin{equation} \label{eq_4_1}
    ACC = \frac{TP + TN}{TP + TN + FP + FN} \times 100
\end{equation}

$ACC$ is the accuracy of the model. While the accuracy computation of the model is a straightforward quantitative representation of the model performance, other metrics have also been proposed to illustrate model performance. 

\begin{equation} \label{eq_4_2}
    TPR = \frac{TP}{TP + FN}
\end{equation}

\begin{equation} \label{eq_4_3}
    FPR = \frac{FP}{FP + TN}
\end{equation}

\begin{figure}[t]
    \hspace*{-0.5cm}
    \includegraphics[width=0.5\textwidth]{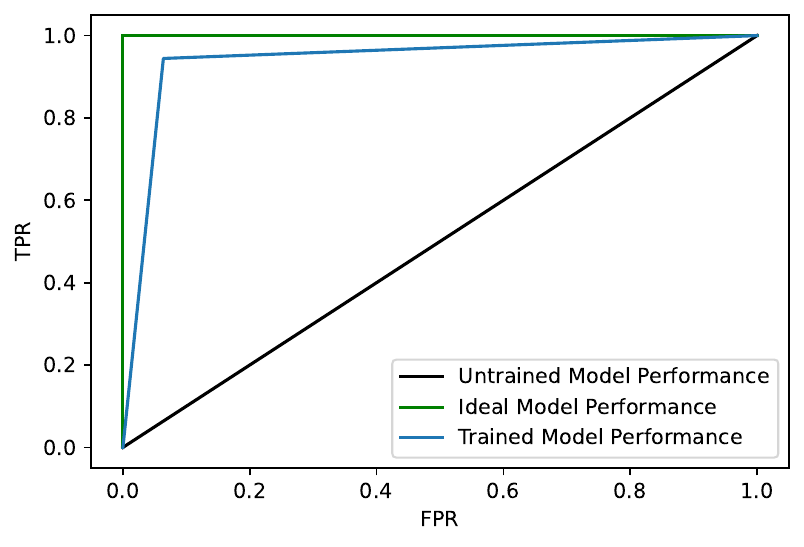}
    \caption{ROC curve of the model showing the performance of the model on the test data}
    \label{fig_4_2}
\end{figure}

$TPR \in (0, 1)$, and $FPR \in (0, 1)$ respectively shown in Eq (\ref{eq_4_2}), and (\ref{eq_4_3}) are defined as the true positive rate and the false positive rate that indicate the performance of the ML model. Ideally, the model $TPR$ should approach one while the model $FPR$ approaches zero. These metrics are shown in a single plot known as the Receiver Operating Characteristic ($ROC$) curve where the model evaluation results could be visualized. Figure \ref{fig_4_2} depicts the $ROC$ curve of the RFC model trained in this study. As the figure shows, the $ROC$ curve of the actual model is close to the ideal case indicating that the trained model performs reasonably. 

\begin{table}[t]
    \centering
    \caption{Different quantitative metrics evaluating the performance of the trained model on the test data}
    \setlength{\tabcolsep}{20pt}
    \begin{tabular}{cccc}
    \hline
      $ACC$   & $TPR$ & $FPR$ & $MCC$  \\
      \hline
      $92.316\%$ & $0.945$ & $0.063$ & $0.823$ \\
      \hline
    \end{tabular}
    \label{tab_4_1}
\end{table}

Although the previously defined metrics quantitatively determined the model performance, since the number of datasets with different classes is unbalanced, another quantitative metric was used as well. Hence, Matthews Correlation Coefficient ($MCC$) was used to evaluate model performance, where $MCC \in [-1, 1]$ \cite{MCC}. The $MCC$ criterion is defined as shown in Eq(\ref{eq_4_4}). The model performance is expected to improve as the $MCC$ criterion approaches the upper limit of its domain. 

\begin{equation} \label{eq_4_4}
    MCC = \frac{TP.TN - FP.FN}{\sqrt{(TP + FN)(TP + FP)(TN + FN)(TN + FP)}}
\end{equation}

As shown in Table \ref{tab_4_1}, the $MCC$ factor is 0.895, thus demonstrating that the model exhibits an acceptable performance on the test data. 

\subsection{Results of the XAI: Shapley Values}

\begin{figure}[t]
    \hspace*{-0.5cm}
    \includegraphics[width=0.5\textwidth]{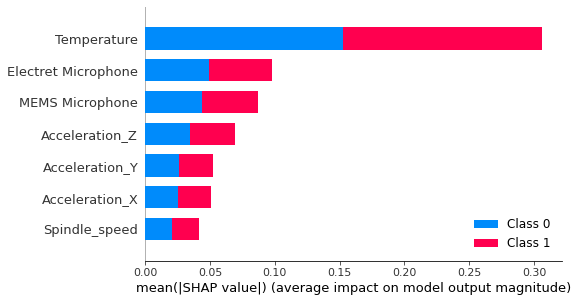}
    \caption{Results of the Shapley criterion implemented on the trained model determining the effect of each feature on the overall decision of the ML algorithm. Class 0 refers to an unworn tool, whereas class 1 indicates the worn tool}
    \label{fig_4_3}
\end{figure}

Figure \ref{fig_4_3} shows the results of implementing the Shapley criterion. As Fig. \ref{fig_4_3} illustrates, temperature is the most significant feature among all the input features followed by the data acquired from the microphones for the decision making by the model for both of the classes. On the contrary, Fig. \ref{fig_4_3} suggests that among all the input features, spindle speed was the least decisive feature cooperating in the final decision of the model. To better understand the effect of each input feature on the model performance, the Shapley criterion was implemented on the individual data points where the model made a correct predictions as well as where the model made a false prediction.

\begin{figure}[h]
    \hspace*{-0.25cm}
     \subfigure[Case I: The model has correctly predicted the state of the tool indicating that the tool is worn]{\includegraphics[width=0.5\textwidth]{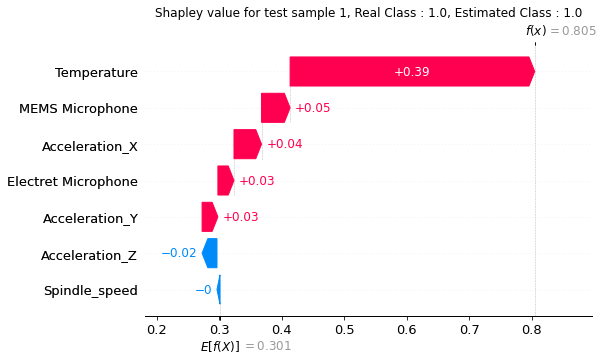}}
    \hfill
    \hspace*{-0.25cm}
    \subfigure[Case II: The model has correctly predicted the state of the tool indicating that the tool is unworn]{\includegraphics[width=0.5\textwidth]{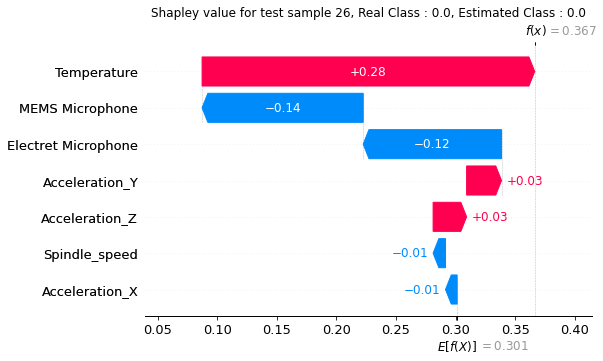}}
    \hfill
    \caption{Shapley values for the cases where the model made true predictions}
    \label{fig_4_4}
\end{figure}

Figure \ref{fig_4_4} shows the Waterfall plots of the Shapley criterion implementation on two instances where the model made a correct prediction. Figure \ref{fig_4_4}(a) represents the Shapley waterfall values for a correct model prediction where the tool was worn. As Fig. \ref{fig_4_4}(a) illustrates, temperature was the most crucial input feature for the model prediction followed by the microphone data. Similarly, Fig. \ref{fig_4_4}(b) shows Shapley values for the case where the model had correctly predicted that the tool was unworn. Similar to the previous case, the figure suggests that the temperature data was the most vital feature for the final model prediction. It could also be interpreted that the spindle speed had very little contribution to the final decision making of the model confirming that this input feature is not a prominent factor for a correct classification. 

\begin{figure}[t]
     \hspace*{-0.25cm}
     \subfigure[Case I: The real class of the input in this case is 0, meaning that the tool is unworn, whereas, the ML model made a false prediction]{\includegraphics[width=0.5\textwidth]{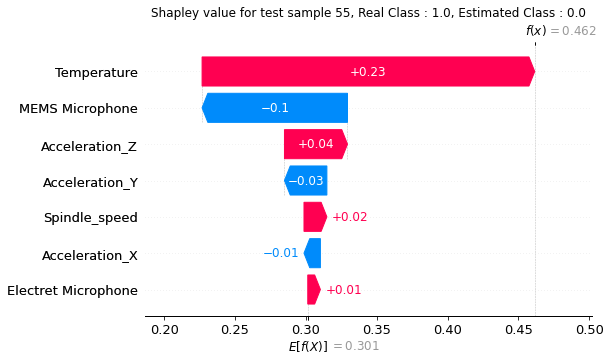}}
    \hfill
    \hspace*{-0.25cm}
    \subfigure[Case II: The real class of the input in this case is 1, meaning that the tool is worn, whereas, the ML model made a false prediction ]{\includegraphics[width=0.5\textwidth]{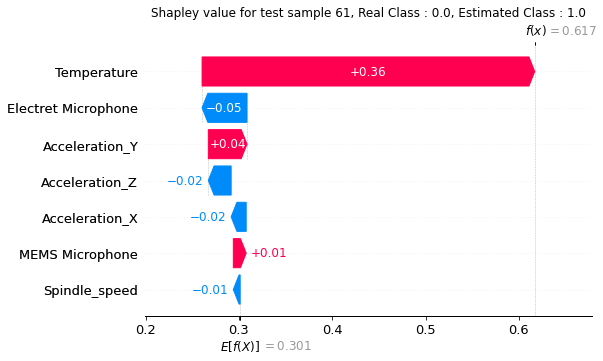}}
    \hfill
    \caption{Shapley values for the cases where the model made false predictions}
    \label{fig_4_5}
\end{figure}

Similar to what was executed on the true prediction cases, the Shapley criterion was implemented on a set of instances where the model incorrectly predicted the state of the tool. The Waterfall plots illustrating the Shapley values of these cases are shown in Fig. \ref{fig_4_5}. In the upper plot, the real class of the instance was 0 meaning that the tool was unworn, whereas, the model had mistakenly predicted the state of the tool to be worn. Contrarily, for the second case, the model predicted that the tool was unworn however, given the input features of the instance, the real class was 1 meaning that the tool was worn. It could be implied from the figure that the temperature was the most important factor in the decision making process of the model. Overall, the developed XAI framework was successfully demonstrated to describe the trained ML model for the task of tool wear prediction. The information provided by XAI framework for this research task is crucial by providing useful information on how the ML model determines the tool state. hence, the information extracted from the XAI framework is critical for manufacturers and operators to estimate the state of the cutting tools used during the production process. In cases where the model makes an incorrect decision, even though the model prediction was incorrect, Shapley criterion indicated that temperature contributed the most in the final decision making of the ML model showing that the temperature was the most vital input feature in determination of the state of the tool. Therefore, the extracted information from the XAI in the false prediction cases could still be helpful for the operators during the manufacturing process.

\section{CONCLUSION}
This research aimed to develop an XAI framework to be utilized to determine the tool wear for manufacturing processes. For this purpose, a set of experiments were carried out using an uncoated tungsten carbide tool to cut a 1018 steel tube on the lathe, and acceleration, acoustics, and temperature data were recorded using different sensors and transducers mounted on the cutting machine for data acquisition. After the data were acquired and preprocessed, a RFC was trained for the task of binary classification on a subset of training datasets. After the training was completed, the ML model performance was evaluated using test datasets illustrating that the ML model had an overall performance accuracy of $92.31\% $. Furthermore, the Shapley criterion was applied to determine to what extend each input feature would be important for the final decision making of the trained RFC model. The results of the research showed that using the tabular representation of Shapley criterion, the framework could explain the decision processes of a complex ML model for the tool wear prediction. This research will enable manufacturers and process engineers to better understand the condition of the cutting tools with reasonable explanations on the model performance. 

Beside its promising results, the current research had several disadvantages regarding the input features selected for the tool wear detection. {The issue was more apparent in cases where the ML model made an incorrect prediction since some features had significant impacts on the prediction of the model affecting the final outcome of the ML model, while other input features did not contribute to the prediction process of the ML model with the same impact. To address this, more thermo-mechanical features will be considered in future works. For instance, the feed rate can be considered as one of the input features for the training process of the ML model and the network explainability will be assessed regarding the new features to determine whether the network could make correct decisions based on the important thermo-mechanical features.} Moreover, the current work heavily relied on the prior knowledge provided by the human expert to do the perception task, more specifically, the ML model in this task is not capable of generalizing its knowledge for a more detailed classification and could only perceive by relying on the prior provided knowledge by the human operator. Therefore, future research will focus on developing an image-based XAI framework using Convolutional Neural Networks (CNN) for better explainability. Additionally, subsequent research will focus on machine-centric perception utilizing meta-learning algorithms where the ML framework is capable of generalizing its' prior knowledge for a detailed and more accurate classification. 

\bibliographystyle{asmems4}

\begin{acknowledgment}
The authors declare they have no conflict of interests that could have appeared to influence the work reported in this paper.
\end{acknowledgment}

\bibliography{asme2e} 

\end{document}